\documentclass[10pt,twocolumn,letterpaper]{article}

\usepackage{iccv}
\usepackage{times}
\usepackage{epsfig}
\usepackage{graphicx}
\usepackage{amsmath}
\usepackage{amssymb}
\usepackage{textcomp}
\usepackage{multirow}
\usepackage{adjustbox}

\usepackage{multirow}
\usepackage{colortbl}
\usepackage{hhline}
\usepackage{subcaption}
\usepackage{enumitem}
\usepackage{verbatim}
\usepackage{tabularx}

\usepackage{gensymb}
\usepackage{flushend}
\usepackage{xcolor}

\usepackage[pagebackref=true,breaklinks=true,letterpaper=true,colorlinks,bookmarks=false]{hyperref}

\iccvfinalcopy 

\def\iccvPaperID{0630} 

\ificcvfinal\fi

\begin{document}

\title{WoodScape: A multi-task, multi-camera fisheye dataset for autonomous driving}

\author{%
Senthil Yogamani, Ciar\'{a}n Hughes, Jonathan Horgan, Ganesh Sistu, Padraig Varley, Derek O'Dea, \\ Michal U\v{r}i\v{c}\'{a}\v{r}, Stefan Milz, Martin Simon, Karl Amende, Christian Witt, Hazem Rashed, \\ Sumanth Chennupati, Sanjaya Nayak, Saquib Mansoor,  
Xavier Perrotton,  Patrick P\'erez\\
{\tt\small \url{https://github.com/valeoai/WoodScape} \hspace{3.5cm} firstname.lastname@valeo.com } \\
}

\twocolumn[{
	\renewcommand\twocolumn[1][]{#1}
	\maketitle
	\begin{center}
		\vspace{-0.6cm}
		\includegraphics[width=\textwidth]{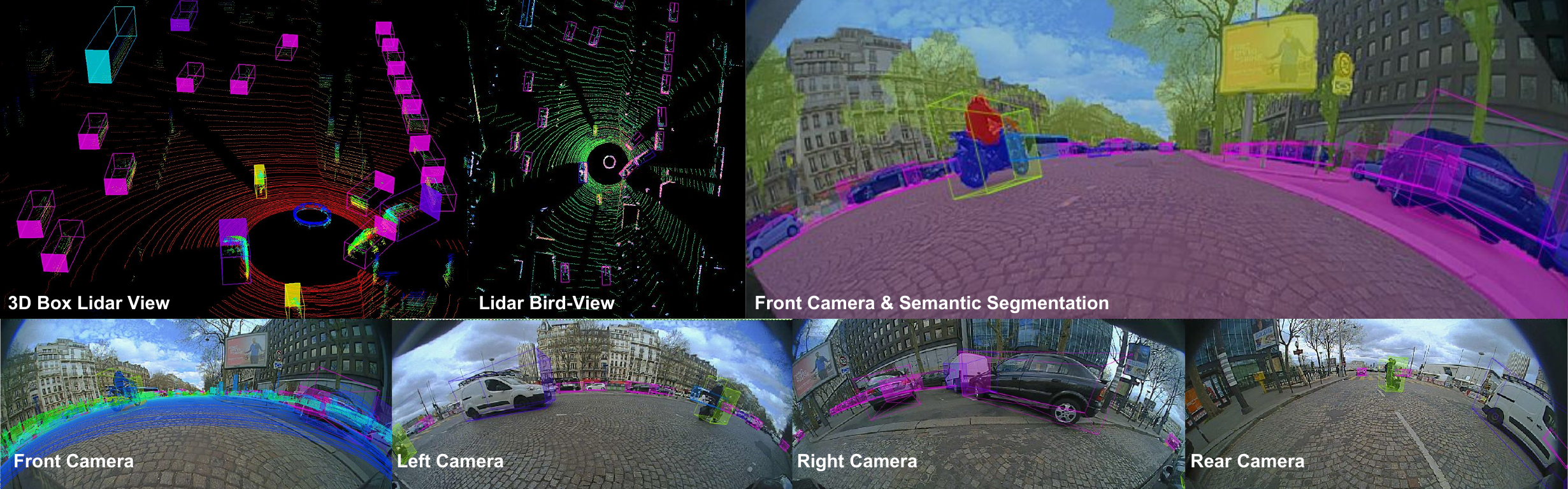}
        \vspace{-0.65cm}
		\captionof{figure}
		{
			We introduce WoodScape, the first fisheye image dataset dedicated to autonomous driving. It contains four cameras covering $360^\degree$  accompanied by a HD laser scanner, IMU and GNSS. Annotations are made available for nine tasks, notably 3D object detection, depth estimation (overlaid on front camera) and semantic segmentation as illustrated here.  
		}
		\label{fig:teaser}
	\end{center}
}]



\begin{abstract}
Fisheye cameras are commonly employed for obtaining a large field of view in surveillance, augmented reality and in particular automotive applications. In spite of their prevalence, there are few public datasets for detailed evaluation of computer vision algorithms on fisheye images. We release the first extensive fisheye automotive dataset, WoodScape, named after Robert Wood who invented the fisheye camera in 1906. WoodScape comprises of four surround view cameras and nine tasks including segmentation, depth estimation, 3D bounding box detection and soiling detection. Semantic annotation of 40 classes at the instance level is provided for over 10,000 images and annotation for other tasks are provided for over 100,000 images. With WoodScape, we would like to encourage the community to adapt computer vision models for fisheye camera instead of using na\"{i}ve rectification. 
\end{abstract}
\vspace{-0.4cm}


\section{Introduction} \label{sec:introduction}

Fisheye lenses provide a large field of view (FOV) using a highly non-linear mapping instead of the standard perspective projection. However, it comes at the cost of strong radial distortion. Fisheye cameras are so-named because they relate to the $180^\degree$ view of the world that a fish has observing the water surface from below, a phenomenon known as Snell's window. Robert Wood originally coined the term in 1906 \cite{wood1906xxiii}, and constructed a basic fisheye camera by taking a pin-hole camera and filling it with water. It was later replaced with a hemispherical lens \cite{bond1922lxxxix}. To pay homage to the original inventor and coiner of the term ``fisheye'', we have named our dataset WoodScape. 

Large FOV cameras are necessary for various computer vision application domains, including video surveillance \cite{kim2016fisheye} and augmented reality \cite{schmalstieg2016augmented}, and have been of particular interest in autonomous driving \cite{horgan2015vision}. In automotive, rear-view fisheye cameras are commonly deployed in existing vehicles for dashboard viewing and reverse parking. While commercial autonomous driving systems typically make use of narrow FOV forward facing cameras at present, full $360^\degree$ perception is now investigated for handling more complex use cases. In spite of this growing interest, there is relatively little literature and datasets available. Some examples of the few datasets that have fisheye are: Visual SLAM ground truth for indoor scenes with omni-directional cameras in \cite{caruso2015large}, SphereNet \cite{coors2018spherenet} containing 1200 labelled images of parked cars using $360^\degree$ cameras (not strictly fisheye) and, in automotive, the Oxford Robotcar dataset \cite{maddern20171} containing a large scale relocalization dataset. \\



\begin{figure}[tb]
\centering
\includegraphics[width=\columnwidth]{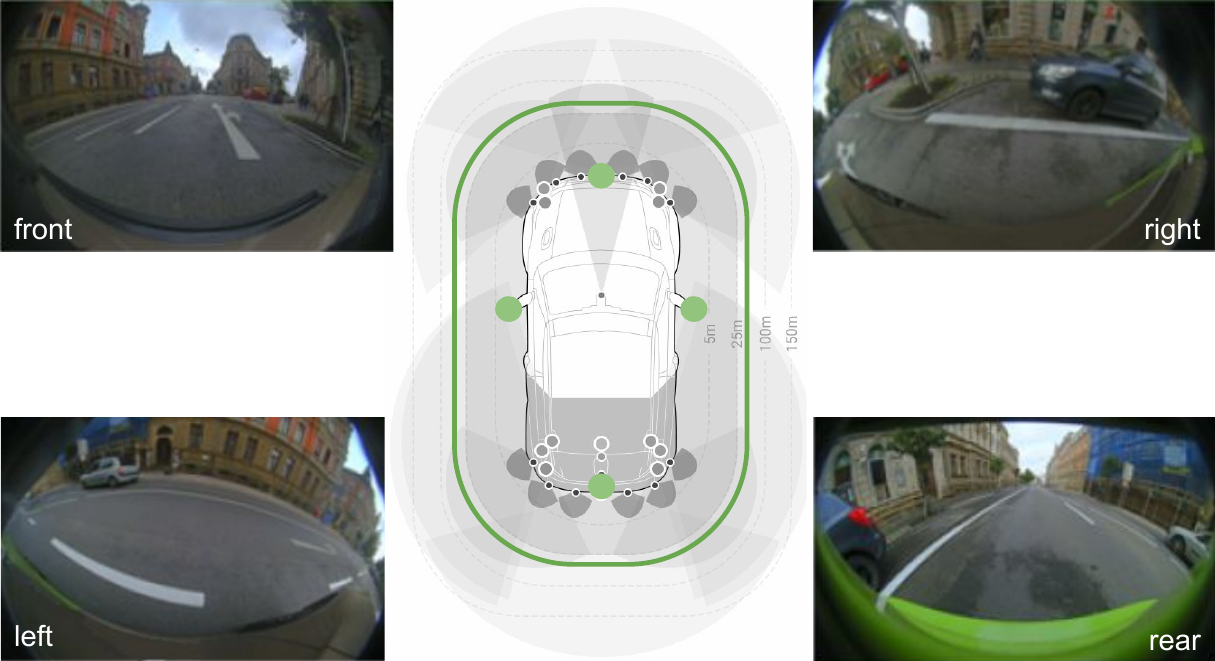}
\vspace{-0.75cm}
\caption{Sample images from the surround-view camera network showing wide field of view and 360$^\circ$ coverage.} 
\vspace{-0.5cm}
\label{fig:svs}
\end{figure}

WoodScape is a comprehensive dataset for $360^\degree$ sensing around a vehicle using the four fisheye cameras shown in Figure \ref{fig:svs}. It aims at complementing the range of already existing automotive datasets where only narrow FOV image data is present: among those, KITTI \cite{geiger2013vision} was the first pioneering dataset with a variety of tasks, which drove a lot of research for autonomous driving; Cityscapes \cite{cordts2016cityscapes} provided the first comprehensive semantic segmentation dataset and Mapillary \cite{neuhold2017mapillary} provided a significantly larger dataset;  Apolloscape \cite{huang2018apolloscape} and BDD100k \cite{yu2018bdd100k} are more recent datasets that push the annotation scale further. WoodScape is unique in that it provides fisheye image data, along with a comprehensive range of annotation types.  A comparative summary of these different datasets is provided in Table \ref{table:datasets}. 
\noindent The main contributions of WoodScape are as follows: 
\begin{enumerate} [nolistsep]
    \item First fisheye dataset comprising of over 10,000 images containing instance level semantic annotation.
    \item Four-camera nine-task dataset designed to encourage unified multi-task and multi-camera models.
    \item Introduction of a novel soiling detection task and release of first dataset of its kind.
    \item Proposal of an efficient metric for the 3D box detection task which improves training time by 95x. \\
\end{enumerate} 

The paper is organized as follows. Section \ref{sec:fisheye} provides an overview of fisheye camera model, undistortion methods and fisheye adaption of vision algorithms. Section \ref{sec:dataset} discusses the details of the dataset including goals, capture infrastructure and dataset design. Section \ref{sec:tasks} presents the list of supported tasks and baseline experiments. Finally, Section \ref{sec:conclusions} summarizes and concludes the paper.


\section{Overview of Fisheye Camera Projections} \label{sec:fisheye}

Fisheye cameras offer a distinct advantage for automotive applications. Given their extremely wide field of view, they can observe the full surrounding of a vehicle with a minimal number of sensors, with just four cameras typically being required for full 360$^\circ$ coverage (Figure \ref{fig:svs}). This advantage comes with some drawbacks in the significantly more complex projection geometry that fisheye cameras exhibit. That is, images from fisheye cameras display severe distortion. 

Typical camera datasets consist of narrow FOV camera data where a simple pinhole projection model is commonly employed. In case of fisheye camera images, it is imperative that the appropriate camera model is well understood either to handle distortion in the algorithm or to warp the image prior to processing. This section is intended to highlight to the reader that the fisheye camera model requires specific attention. We provide a brief overview and references for further details, and discuss the merits of operating on the raw fisheye versus undistortion of the image.

\subsection{Fisheye Camera Models} \label{sec:fisheyemodel}

\begin{figure}[t]
\centering
\includegraphics[width=\columnwidth,trim={0.5cm 0 0.5cm 0.5cm},clip]{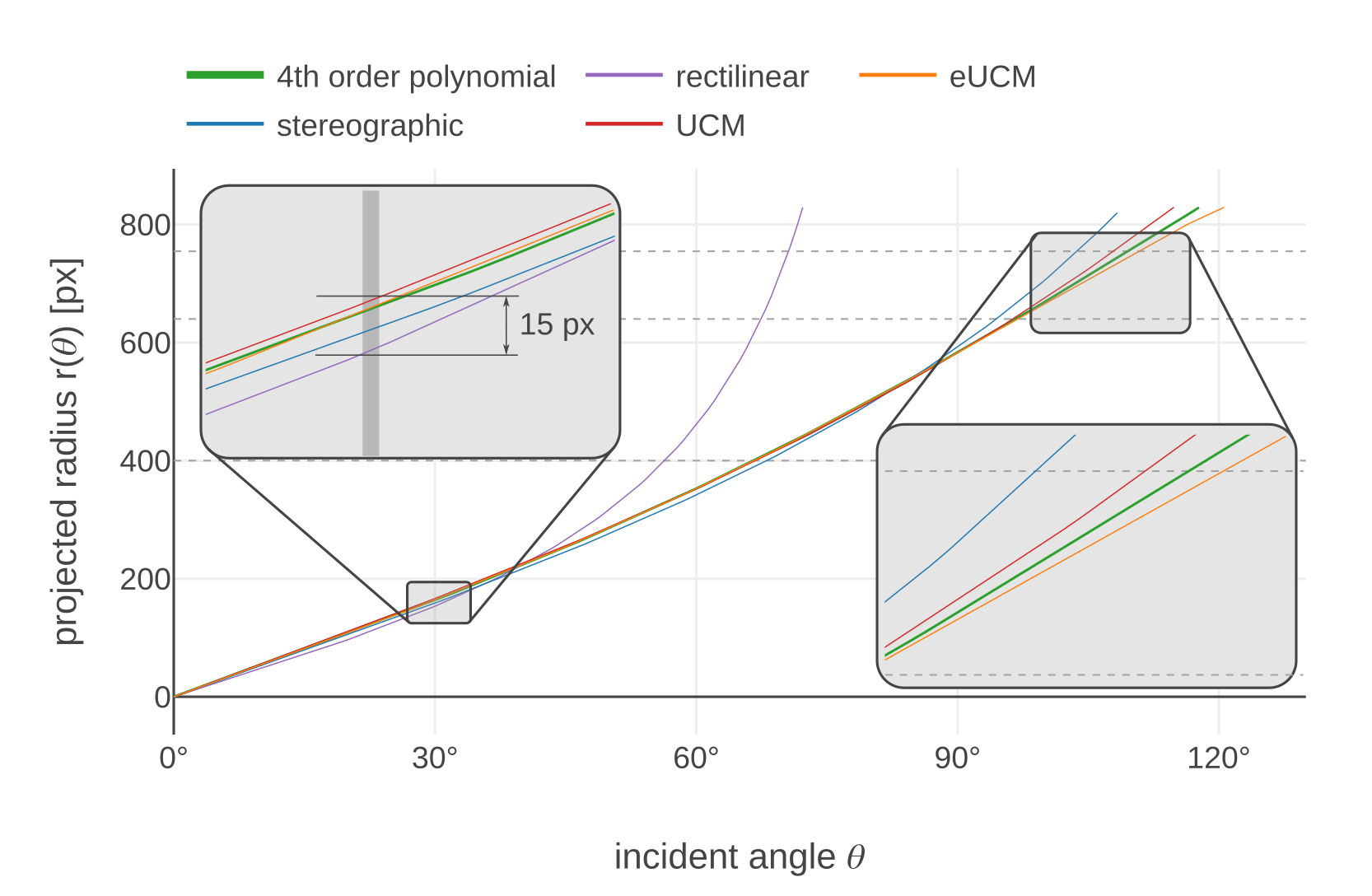}
\vspace{-0.75cm}
\caption{Comparison of fisheye models.}
\vspace{-0.35cm}
\label{fig:fsh}
\end{figure}

Fisheye distortion is modelled by a radial mapping function $r(\theta)$, where $r(\theta)$ is the distance on the image from the centre of distortion, and is a function of the angle $\theta$ of the incident ray against the optical axis of the camera system. The centre of distortion is the intersection of the optical axis with the image plane, and is the origin of the radial mapping function $r(\theta)$. Stereographic projection \cite{herbert1987area} is the simplest model which uses a mapping from a sphere to a plane. More recent projection models are Unified Camera Model (UCM) \cite{barreto2006unified, caruso2015large} and eUCM (Enhanced UCM) \cite{Khomutenko2016eucm}. More detailed analysis of accuracy of various projection models is discussed in \cite{hughes2010fisheye}. These models are not a perfect fit for fisheye cameras as they encode a specific geometry (e.g. spherical projection), and errors arising in the model are compensated by using an added distortion correction component. 

In WoodScape, we provide model parameters for a more generic fisheye intrinsic calibration that is independent of any specific projection model, and does not require the added step of distortion correction. Our model is based on a fourth order polynomial mapping incident angle to image radius in pixels ($r(\theta) = a_1 \theta + a_2 \theta^2 + a_3 \theta^3 + a_4 \theta^4$). 
In our experience, higher orders provide no additional accuracy. Each video sequence in the dataset is provided with parameters for the fourth order polynomial model of fisheye intrinsics.

As a comparison, to give the reader an understanding of how different models behave, Figure \ref{fig:fsh} shows the mapping function $r(\theta)$ for five different projection models, which are Polynomial, Rectilinear, Stereographic, UCM and eUCM.
The parameters of the fourth order polynomial are taken from a calibration of our fisheye lens. We optimized the parameters for the other models to match this model in a range of 0$^\circ$ to 120$^\circ$ (i.e. up to FOV of $240^\circ$). The plot indicates that the difference to the original fourth order polynomial is about four pixels for UCM and one pixel for eUCM for low incident angles. For larger incident angles, these models are less precise.



\begin{figure}[tb]
    \centering
    \includegraphics[width=0.75\columnwidth]{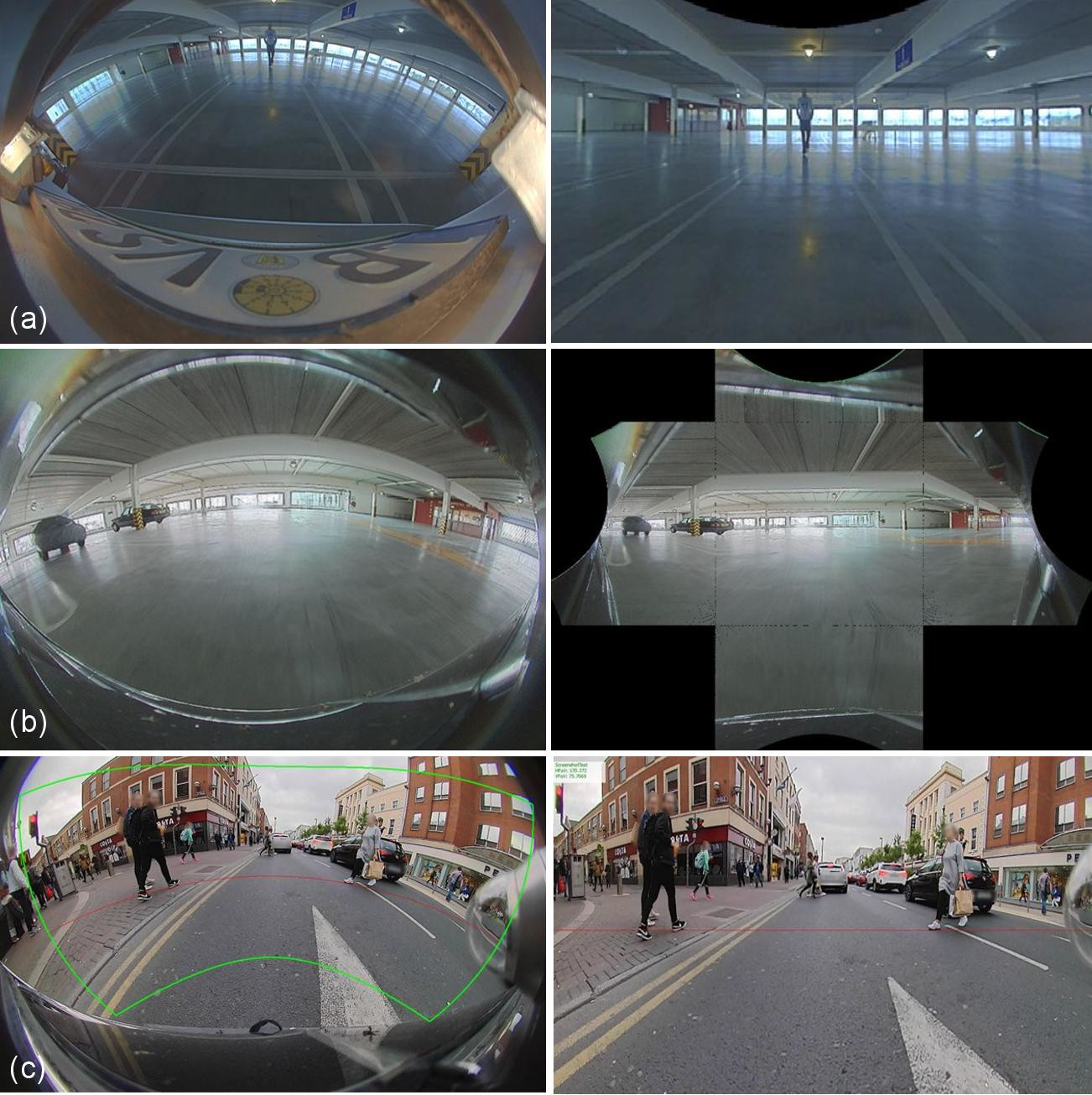}
    \vspace{-0.15cm}
    \caption{Undistorting the fisheye image: 
    (a) Rectilinear correction; (b) Piecewise linear correction; (c) Cylindrical correction. Left: raw image; Right: undistorted image.}\label{fig:projections}
    \vspace{-0.5cm}
\end{figure}



\subsection{Image Undistortion vs. Model Adaptation} \label{sec:linearization}

Standard computer vision models do not generalize easily to fisheye cameras because of large non-linear distortion. For example, translation invariance is lost for a standard convolutional neural net (CNN). The na\"{i}ve way to develop algorithms for fisheye cameras is to perform rectilinear correction so that standard models can be applied. The simplest undistortion is to re-warp pixels to a rectilinear image as shown in Figure \ref{fig:projections} (a). But there are two major issues. Firstly, the FOV is greater than 180$^\circ$, hence there are rays incident from behind the camera and it is not possible to establish a complete mapping to a rectilinear viewport. This leads to a loss of FOV, this is seen via the missing yellow pillars in the corrected image. Secondly, there is an issue of resampling distortion, which is more pronounced near the periphery of the image where a smaller region gets mapped to a larger region. 

The missing FOV can be resolved by multiple linear viewports as shown in Figure \ref{fig:projections} (b). However there are issues in the transition region from one plane to another. This can be viewed as a piecewise linear approximation of the fisheye lens manifold. Figure \ref{fig:projections} (c) demonstrates a quasi-linear correction using a cylindrical viewport, where it is linear in vertical direction and straight vertical objects like pedestrians are preserved. However, there is a quadratic distortion along the horizontal axis. In many scenarios, it provides a reasonable trade-off but it still has limitations. In case of learning algorithms, a parametric transform can be optimized for optimal performance of the target application accuracy.

Because of fundamental limitations of undistortion, an alternate approach of adapting the algorithm incorporating fisheye model discussed in previous section could be an optimal solution. In case of classical geometric algorithms, an analytical version of non-linear projection can be incorporated. For example, Kukelova et al.~\cite{kukelova2015radial} extend homography estimation by incorporating radial distortion model. In case of deep learning algorithms, a possible solution could be to train the CNN model to learn the distortion. However, the translation invariance assumption of CNN fundamentally breaks down due to spatially variant distortion and thus it is not efficient to let the network learn it implicitly. This had led to several adaptations of CNN to handle spherical images such as \cite{su2018kernel} and \cite{coors2018spherenet}. However, spherical models do not provide an accurate fit for fisheye lenses and it is an open problem.


\section{Overview of WoodScape Dataset} \label{sec:dataset}

\subsection{High-Level Goals}

\textbf{Fisheye:} One of the main goals of this dataset is to encourage the research community to develop vision algorithms natively on fisheye images without undistortion. There are very few public fisheye datasets and none of them provide semantic segmentation annotation. Fisheye is particularly beneficial to automotive low speed manoeuvring scenarios such as parking \cite{heimberger2017computer} where accurate full coverage near field sensing can be achieved with just four cameras.

\textbf{Multi-camera:} Surround view systems have at least four cameras rigidly connected to the body of the car. Pless \cite{pless2003using} did pioneering work in deriving a framework for modeling a network of cameras as one, this approach is useful for geometric vision algorithms like visual odometry. However, for semantic segmentation algorithms, there is no literature on joint modeling of rigidly connected cameras. 

\textbf{Multi-task:} Autonomous driving has various vision tasks and most of the work has been focused on solving individual tasks independently. However, there is a recent trend \cite{8100062,8500504,sistu2019neurall,visapp19} to solve tasks using a single multi-task model to enable efficient reuse of encoder features and also provide regularization while learning multiple tasks. However, in these cases, only the encoder is shared and there is no synergy among decoders. Existing datasets are primarily designed to facilitate task-specific learning and they don't provide simultaneous annotation for all the tasks. 
We have designed our dataset so that simultaneous annotation is provided for various tasks with some exceptions due to practical limitations of optimal dataset design for each task. 



\begin{figure}[tb]
    \centering
    \includegraphics[width=\columnwidth]{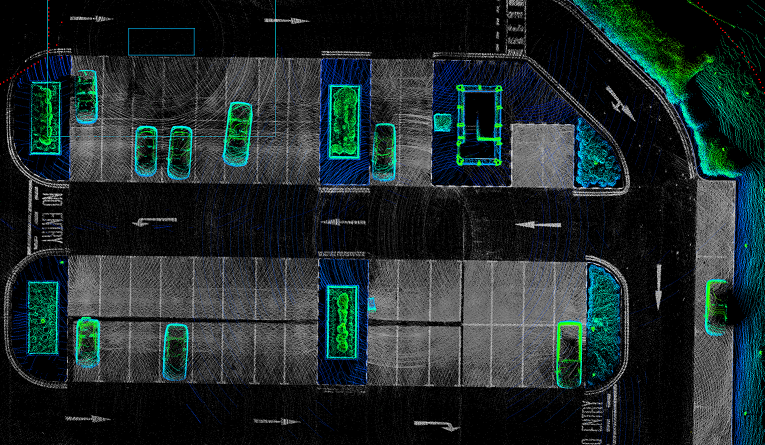}
    \vspace{-0.7cm}
    \caption{SLAM point cloud top-view of a parking lot. Height of the objects is color coded (green for high value, blue for medium value and grayscape for low value).}\label{fig:Velodyne}
    \vspace{-0.5cm}
\end{figure}

\subsection{Dataset Acquisition}
Our diverse dataset originates from three distinct geographical locations: USA, Europe, and China. While the majority of data was obtained from saloon vehicles there is a significant subset from a sports utility vehicle ensuring a strong mix in sensor mechanical configurations. Driving scenarios are divided across the  highway, urban driving and parking use cases. 
Intrinsic and extrinsic calibrations are provided for all sensors as well as timestamp files to allow synchronization of the data. Relevant vehicle's mechanical data (e.g. wheel circumference, wheel base) are included. High-quality data is ensured via quality checks at all stages of the data collection process. 
Annotation data undergoes a rigorous quality assurance by highly skilled reviewers.
The sensors recorded for this dataset are listed below:
\begin{itemize} [nosep]
  \item 4x 1MPx RGB fisheye cameras ($190^\circ$ horizontal FOV)
  \item 1x LiDAR rotating at 20Hz (Velodyne HDL-64E)
  \item 1x GNSS/IMU (NovAtel Propak6 \& SPAN-IGM-A1)
  \item 1x GNSS Positioning with SPS (Garmin 18x)
  \item Odometry signals from the vehicle bus. \\
\end{itemize}



\begin{table*}[htb]
\centering

\caption{Summary of various autonomous driving datasets containing semantic annotation} \label{table:datasets}
\vspace{-0.25cm}
\arrayrulecolor{black}
\begin{adjustbox}{width=\textwidth}


\begin{tabular}{llccccccc}
\hline
\textbf{Task/Info}                       & \textbf{Quantity} & \begin{tabular}{@{}c@{}}\textbf{KITTI} \\ \cite{geiger2013vision} \end{tabular}                                               &  \begin{tabular}{@{}c@{}}\textbf{Cityscapes} \\ \cite{cordts2016cityscapes} \end{tabular}                                   & \begin{tabular}{@{}c@{}}\textbf{Mapillary} \\ \cite{neuhold2017mapillary} \end{tabular}  & \begin{tabular}{@{}c@{}}\textbf{nuScenes} \\ \cite{nuScenes}  \end{tabular}                                                           &\begin{tabular}{@{}c@{}}\textbf{ApolloScape} \\ \cite{huang2018apolloscape}  \end{tabular}   & \begin{tabular}{@{}c@{}}\textbf{BDD100k} \\ \cite{yu2018bdd100k}  \end{tabular}  & \begin{tabular}[c]{@{}c@{}} \textbf{WoodScape}\\ \textbf{Ours}\end{tabular}                                                   \\ \hline \hline
\multirow{3}{*}{Capture Information}    & Year             & 2012/14/15                                                                             & 2016                                                                  & 2017               & 2018                                                                           & 2018                 & 2018             & 2018/19                                                               \\ 
                                    & State/cities 
                                    & 1/1                                                               & 2/50                                                                  & 50+/100+           & 2/2                                                                            & 1/4                  & 1/4              & 5+/10+                                                                \\  
                                    & Other sensors          & \begin{tabular}[c]{@{}c@{}}1 LiDAR\\ GPS\end{tabular}      & -                                                                    & -                 & \begin{tabular}[c]{@{}c@{}}1 LiDAR\\ GPS, IMU\\ 5 RADAR\end{tabular}            & \begin{tabular}[c]{@{}c@{}}2 LiDAR\\ GNSS \\ IMU\end{tabular}                    & \begin{tabular}[c]{@{}c@{}}1 GPS\\ IMU\end{tabular}           & \begin{tabular}[c]{@{}c@{}}1 LiDAR\\ GNSS\\ IMU\end{tabular} \\ \hline 
\multirow{3}{*}{Camera Information} & Cameras       & 4                                                                        & 2                                                                    & -                 & 6                                                                              & 6                   & 1                & 4                                                                     \\ 
                                    & Tasks            & 6                                                                                                              & 1                                                          & 1       & 1          & 4                    & 2                & 9                                                                     \\ \hline 
\multirow{2}{*}{Segmentation}       & Classes       & 8                                                                                                                 & 30                                                                    & 66                 & -                                                                             & 25                   & 40               & 40                                                                    \\ 
                                    & Frames   & 400                                                                                             & 5k & 25k         & -                                                                             & 140k                 & 5.7k   & 10k                                                                   \\ \hline 
\multirow{2}{*}{2D Bounding Box$^1$}       & Classes       & 3                                                                                                                     & -                                                                    & -                 & -                                                                             & -                   & 10               & 7                                                                     \\ 
                                    & Frames   & 15k                                                                                                            & -                                                                    & -                 & -                                                                             & -                   & 5.7k  & 10k                                                                   \\ \hline
\multirow{2}{*}{3D Bounding  Box}   & Classes       & 3                                                                                                                     & -                                                                    & -                 & 25                                                                             & 1                    & -               & 3                                                                     \\ 
                                    & Frames   & 15k                                                                                                            & -                                                                    & -                 & 40k                                                                              & 5k+                  & -               & 10k                                                                    \\ \hline 
Depth Estimation                    & Frames   & 93k                                                                                                           & -                                                                    & -                 & -                                                                             & -                   & -               & 400k                                                                  \\ 
Motion Segmentation                 & Frames   & 1.6k                                                                                                               & -                                                                    & -                 & -                                                                             & -                   & -               & 10k                                                                   \\ 
Soiling Detection                   & Frames   & -                                                                                                                    & -                                                                    & -                 & -                                                                             & -                   & -               & 5k                                                                   \\ 
Visual SLAM/Odometry                      & Videos   & 33                                                                                                               & -                                                                    & -                 & -                                                                             & -                   & -               & 50                                                                 \\ 
End-to-end Driving                  & Videos  & -                                                                                                                    & -                                                                    & -                 & -                                                                             & -                   & -               & 500                                                                 \\ 
Synthetic Data                      & Frames    & -                                                                                                                    & -                                                                    & -                 & -                                                                             & -                   & -               & 10k                                                                  \\
\hline 
\multicolumn{9}{l}{\footnotesize $^1$2D box annotation can be obtained for other datasets from instance segmentation.}
\end{tabular}

\arrayrulecolor{black}
\end{adjustbox}
\vspace{-0.5cm}
\end{table*}

Our WoodScape dataset provides labels for several autonomous driving tasks including semantic segmentation, monocular depth estimation, object detection (2D \& 3D bounding boxes), visual odometry, visual SLAM, motion segmentation, soiling detection and end-to-end driving (driving controls). In Table \ref{table:datasets}, we compare several properties of popular datasets against WoodScape. In addition to providing fisheye data, we provide data for many more tasks than is typical (nine in total), providing completely novel tasks such as soiled lens detection. Images are provided at 1MPx 24-bit resolution and videos are uncompressed at 30fps ranging in duration from 30s to 120s. The dataset also provides a set of synthetic data using accurate models of the real cameras, enabling investigations of additional tasks.  
The camera has a HDR sensor with a rolling shutter and a dynamic range of 120 dB. It has features including black level correction, auto-exposure control, auto-gain control, lens shading (optical vignetting) compensation, gamma correction and automatic white balance for color correction. 

\begin{figure}[tb]    
    \centering
    \includegraphics[width=0.8\columnwidth,trim={50 55 50 55},clip]{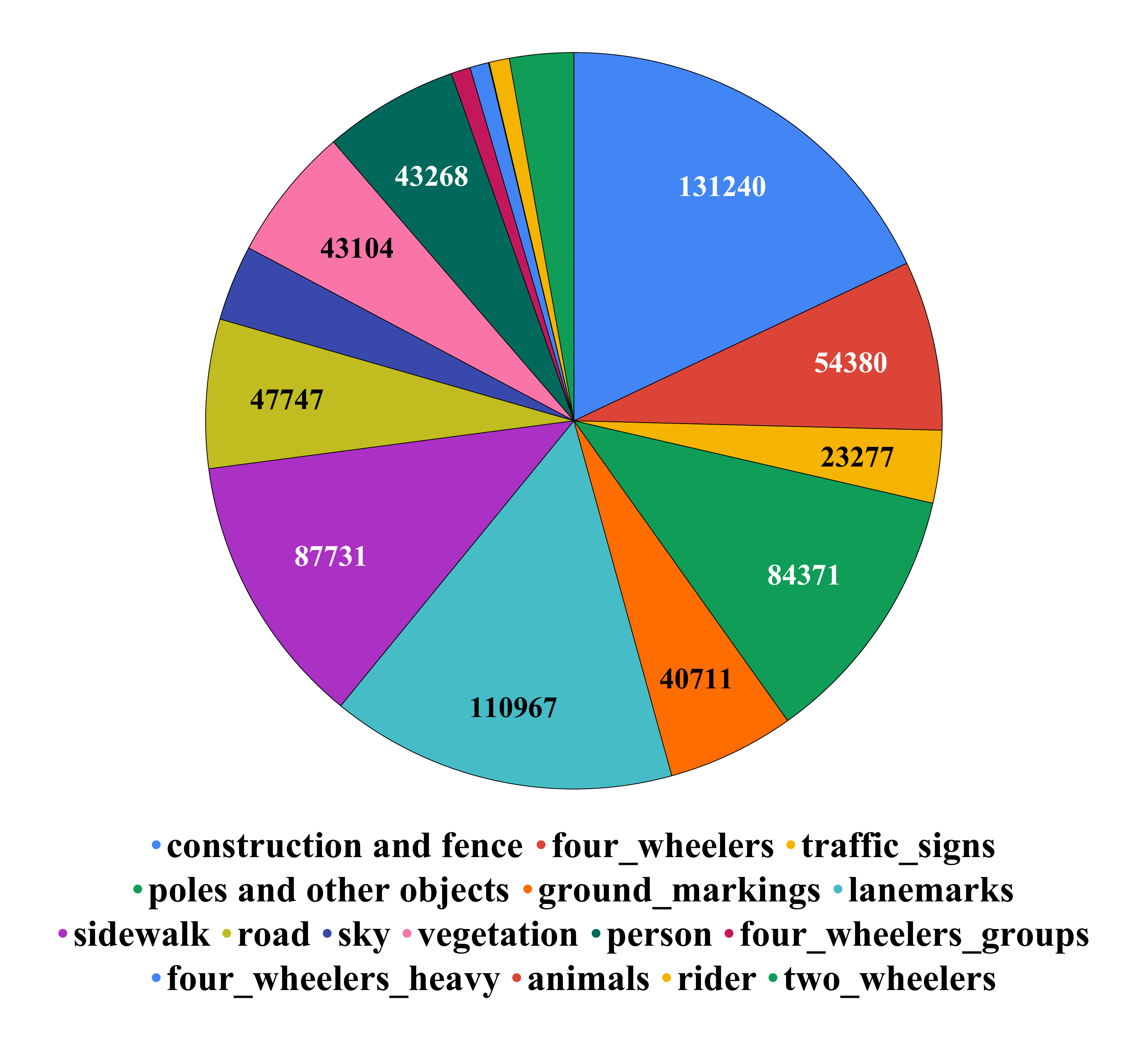}
    \vspace{-0.3cm}
    \caption{Distribution of instances of semantic segmentation classes in WoodScape. Minimum size is 300 pixels.} 
    \label{fig:classDist}
    \vspace{-0.5cm}
\end{figure}

The laser scanner point cloud provided in our data set is accurately preprocessed using a commercial SLAM algorithm to provide a denser point cloud ground truth for tasks such as depth estimation and visual SLAM, as shown in Figure \ref{fig:Velodyne}. In terms of recognition tasks, we provide labels for forty classes, the distribution of the main classes is shown in Figure \ref{fig:classDist}. Note, that for the purposes of display in this paper, we have merged some of the classes in Figure \ref{fig:classDist} (e.g. `two\_wheelers' is a merge of `bicycle' and `motorcycle').

\subsection{Dataset Design}

The design of a dataset for machine learning is a very complex task. Unfortunately, due to the overwhelming success of deep learning, recently it does not get as much attention as it still deserves in our opinion. 
However, at the same time, it was shown that careful inspection of the training sets for outliers improves the robustness of deep neural networks~\cite{Liu-2018}, especially with regards to the adversarial examples. Therefore, we believe that whenever a new dataset is released, there should be a significant effort spent not only on the data acquisition but also on the careful consistency check and on the database splitting for the needs of training, model selection and testing. \\ 

\noindent \textbf{Sampling strategy:} Let us define some notation and naming conventions, which we will refer to first (we follow the definitions provided in~\cite{Borovicka-2012}). A \textit{population} is a set of all existing feature vectors. A subset of the population collected during some process is called a \textit{sample set} $\mathcal{S}$. A \textit{representative set} $\mathcal{S^*}$ is significantly smaller than $\mathcal{S}$, while capturing most of the information from $\mathcal{S}$ (compared to any different subset of the same size), and has low redundancy among the representatives it contains.

In an ideal world, we would like our training set to be equal to $\mathcal{S^*}$. This is extremely difficult to achieve in practice. One approach to approximate this is the concept of the \textit{minimal consistent subset of a training set}, where, given a training set $\mathcal{T}$, we are interested in a subset $\mathcal{T^*}$, being the smallest set such that $\mathrm{Acc}(\mathcal{T^*}) = \mathrm{Acc}(\mathcal{T})$, where $\mathrm{Acc}(\cdot)$ denotes the selected accuracy measure (e.g. the Jaccard index). Note, that computation of accuracy implies having the ground truth labels. The purpose is to reduce the size of the training set by removing non-informative samples, which do not contribute to improving the learned model, and therefore put some ease on the annotation efforts.

There are several ways of obtaining $\mathcal{T^*}$. One frequently used approach is instance selection~\cite{Olvera-Lopez-2010, Liu-2002, Jankowski-2004}. There are two main groups of instance selection: wrappers and filters. The wrapper based methods use a selection criterion based on the constructed classifier's accuracy. Filter based methods, on the other hand, use a selection criterion which is based on an unrelated selection function. 
The concept of a minimal consistent subset is crucial for our setup, where we record image data from video cameras. Collecting frames at a frame rate of $30$fps, particularly at low speeds, ultimately leads to significant image overlap, therefore, having an effective sampling strategy to distill the dataset is critical. 
{We used a combination of a wrapper method using selection criterion based on the classifier's accuracy \cite{Olvera-Lopez-2010} and a simple filter based on the image similarity measurement. 
} \\

\noindent \textbf{Data splitting and class balancing:} 
The dataset is split into three chunks in ratio of $6:1:3$, namely training, validation, and testing. For classical algorithms, all the data can be used for testing. As the names suggest, the training part will serve for training purposes only, the validation part can be either joined with the training set (e.g. when the sought model does not require hyper-parameter selection) or be used for model selection, and finally, the testing set is used for model evaluation purposes only. The dataset supports correct hypothesis evaluation~\cite{Uricar-2019b}, therefore multiple splits are provided ($5$ in total). Depending on the particular task (see Section~\ref{sec:tasks}, for the full list), the class imbalance may be an issue~\cite{Guo-2008}, therefore, task-specific splits are also provided. Full control of the splitting mechanism is provided allowing for each class to be represented equally within each split (i.e. stratified sampling).\\

\noindent \textbf{GDPR challenges:} The recent General Data Protection Regulation (GDPR) regulation in Europe has given rise to challenges in  making our data publicly available. More than one third of our dataset is recorded in Europe and is therefore GDPR sensitive due to visible faces of pedestrians and license plates. 
{There are three primary ways to handle privacy namely (1) Manual blurring, (2) GAN based re-targeting and (3) Stringent data-handling license agreement. Blurring is the commonly used approach wherein privacy sensitive regions in the image are manually blurred. There is also the possibility of using GAN based re-targeting wherein faces are exchanged by automatically generated ones \cite{korshunov2018deepfakes}.
In the recent EuroCity persons dataset \cite{braun2019eurocity}, the authors argued that any anonymization measure will introduce a bias. Thus they released their dataset with original data and a license agreement which enforces the user to strictly adhere to GDPR. We will follow a similar approach. 
}

\section{Tasks, Metrics and Baseline experiments} \label{sec:tasks}

Due to limited space, we briefly describe the metrics and baseline experiments for each task and they are summarized in Table \ref{tab:all-results}. Test dataset for each task consists of 30\% of the respective number of annotated samples listed in Table \ref{table:datasets}. Code is available on WoodScape GitHub and sample video results are shared in supplementary material.

\subsection{Semantic Segmentation}

Semantic segmentation networks for autonomous driving \cite{siam2017deep} have been successfully trained directly on fisheye images in~\cite{deng2017cnn, saez2018cnn}. Due to absence of fisheye datasets, they make use of artificially warped images of Cityscapes for training and testing was performed on fisheye images. However, the artificial images cannot increase the  originally captured FOV. 
Our semantic segmentation dataset provides pixel-wise labels for $40$ object categories, comparatively Cityscapes dataset~\cite{cordts2016cityscapes} provides $30$ for example. Figure \ref{fig:classDist} illustrates the distribution of main classes. We use ENet~\cite{paszke2016enet} to generate our baseline results. We fine-tune their model for our dataset by training with categorical cross entropy loss and Adam~\cite{kingma2014adam} optimizer. We chose Intersection over Union (IoU) metric~\cite{everingham2010pascal} to report the baseline results shown in Table~\ref{tab:all-results}. We acheive a mean IoU of $51.4$ on this test set. Figure \ref{fig:segdet} shows sample results of segmentation on fisheye images from our test set. The four camera images are treated the same, however it would be interesting to explore customization of the model for each camera. The dataset also provides instance segmentation labels to explore panoptic segmentation models~\cite{li2018weakly}.








\subsection{2D Bounding Box Detection}

\begin{figure*}[tb]
    \centering
    \includegraphics[width=.99\textwidth]{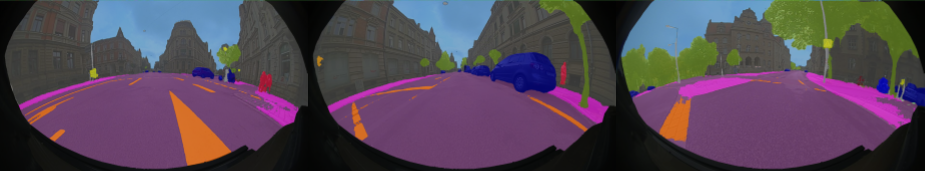}

\begin{subfigure}{.33\textwidth}
    \includegraphics[width=\textwidth,trim={0 2.5cm 0 0},clip]{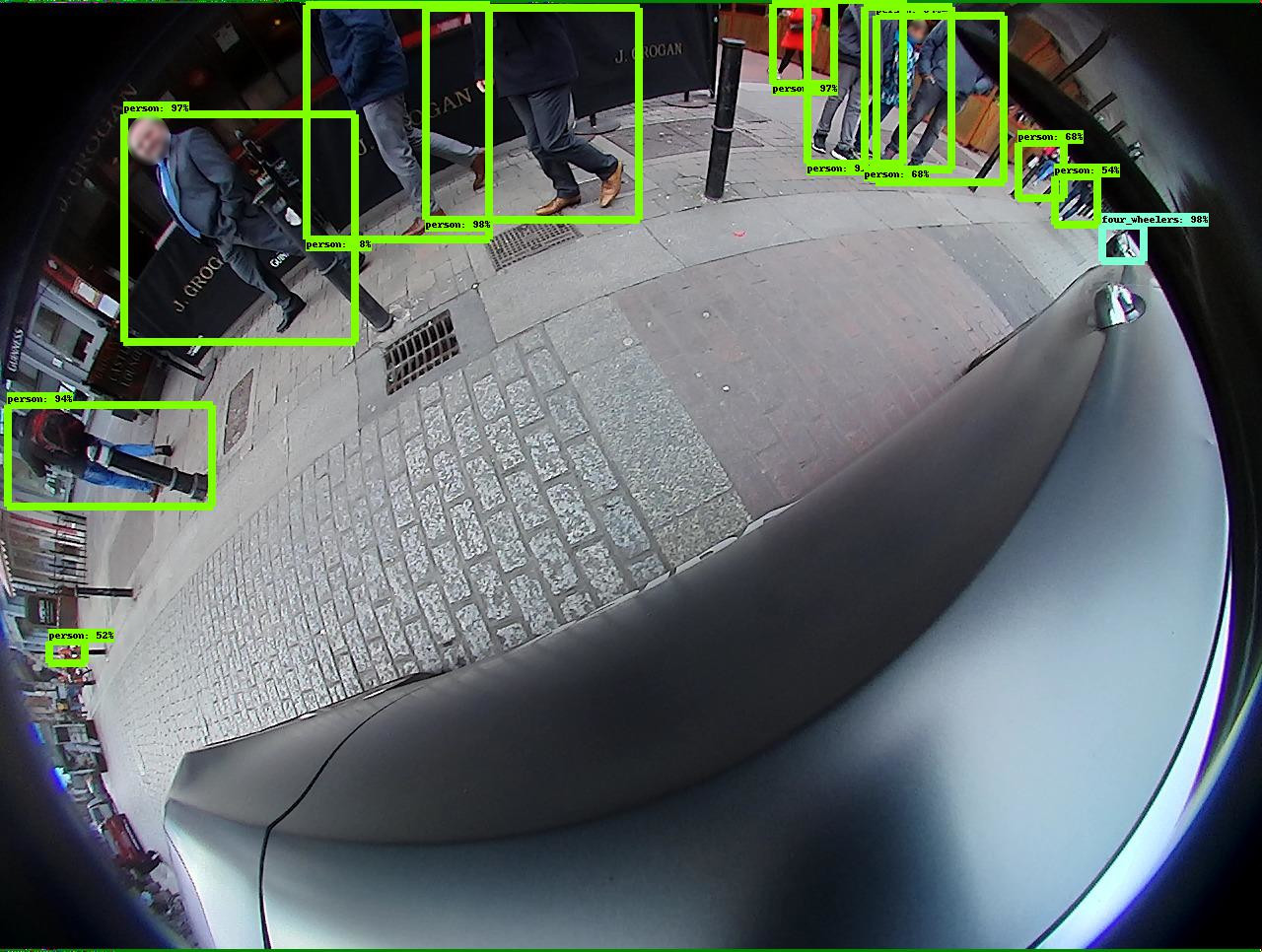}
\end{subfigure}%
\begin{subfigure}{.33\textwidth}
    \includegraphics[width=\textwidth,trim={0 2.5cm 0 0},clip]{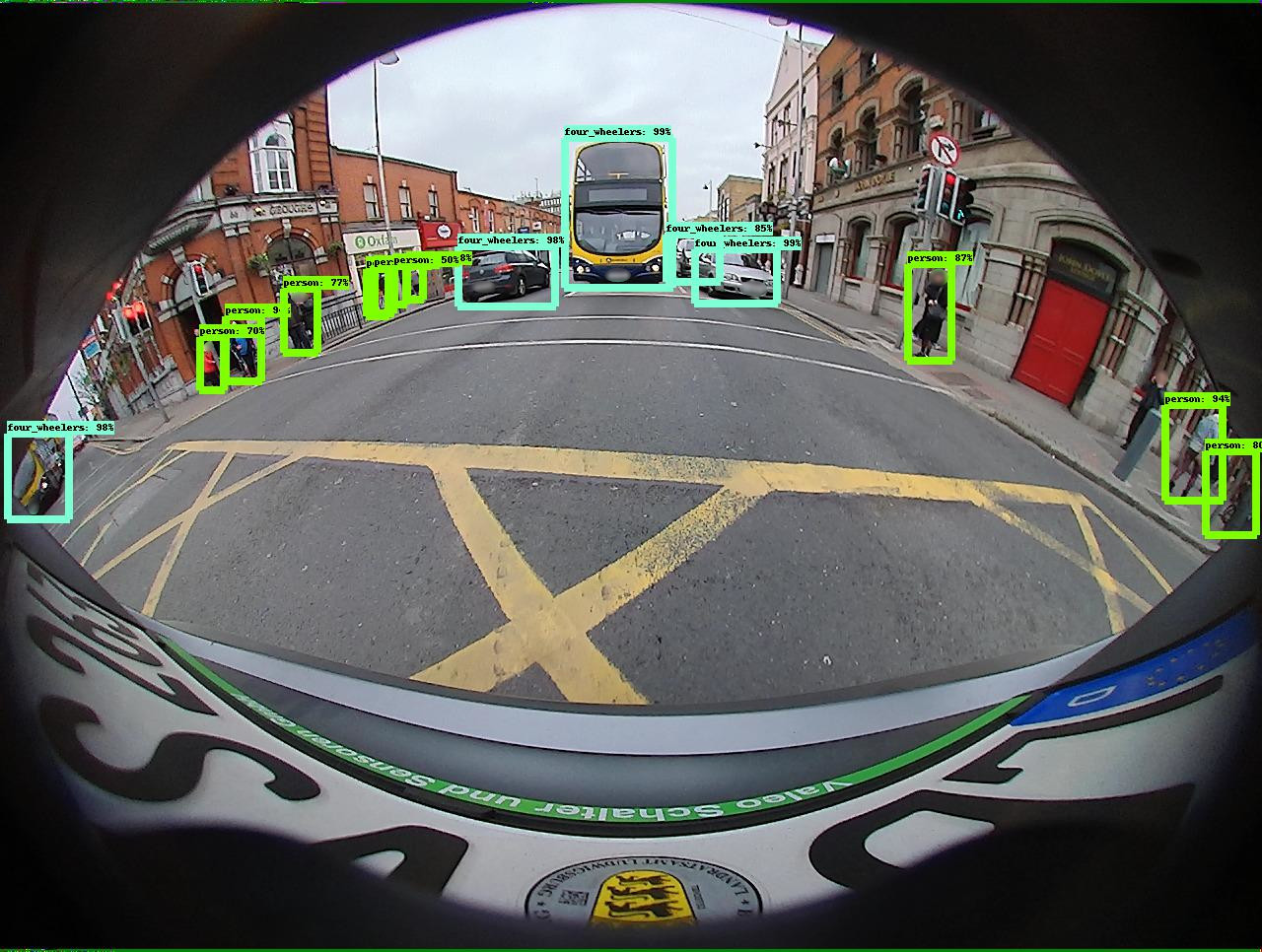}
\end{subfigure}%
\begin{subfigure}{.33\textwidth}
    \includegraphics[width=\textwidth,trim={0 2.5cm 0 0},clip]{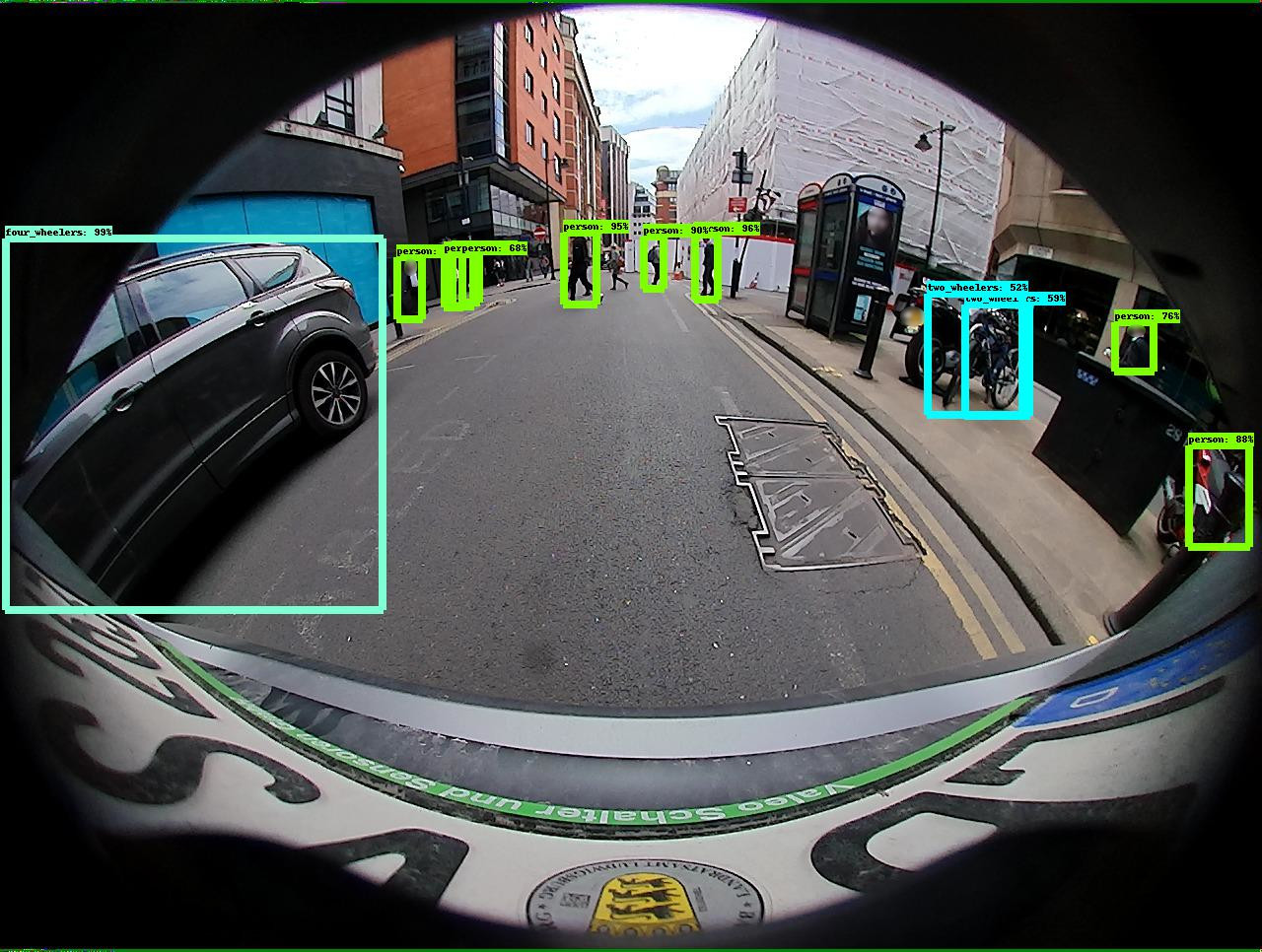}
\end{subfigure}%
\vspace{-0.30cm}
\caption{Qualitative results of Segmentation using ENet \cite{paszke2016enet} (top) and Object detection using Faster RCNN \cite{ren2015faster} (bottom)}
\label{fig:segdet}
\vspace{-0.25cm}
\end{figure*}

Our $2$D object detection dataset is obtained by extracting bounding boxes from instance segmentation labels for $7$ different object categories including pedestrians, vehicles, cyclist and motorcyclist. We use Faster R-CNN~\cite{ren2015faster} with ResNet101~\cite{he2016deep} as encoder. We initialize the network with ImageNet~\cite{deng2009imagenet} pre-trained weights. We fine-tune our detection network by training on both KITTI~\cite{Geiger2012CVPR} and our object detection datasets. Performance of $2$D object detection is reported in terms of mean average precision (mAP) when IoU$\ge0.5$ between predicted and ground truth bounding boxes. We achieve a mAP score of $31$ which is significantly less than the accuracy achieved in other datasets. This was expected as bounding box detection is a difficult task on fisheye (the orientation of objects in the periphery of images being very different from central region). To quantify this better, we tested a pre-trained network for person class, and a poor mAP score of $12$ was achieved compared to our dataset trained value of $45$. Sample results of the fisheye trained model are illustrated in Figure~\ref{fig:segdet}. We observe that it is necessary to incorporate the fisheye geometry explicitly, which is an open research problem.

\subsection{Camera Soiling Detection}

\begin{table}[t]
\centering

\caption{Summary of results of baseline experiments.} 
\vspace{-0.25cm}
\begin{adjustbox}{width=0.5\textwidth}
\begin{tabular}{lccc}
\hline
\textbf{Task}                      & \textbf{Model}                & \textbf{Metric}              & \textbf{Value} \\ \hline \hline
Segmentation                       & ENet \cite{paszke2016enet}                           & IoU                          & 51.4           \\ 
2D Bounding Box                       & Faster R-CNN \cite{ren2015faster}
& mAP (IoU\textgreater{}0.5)   & 31           \\ 
\multirow{1}{*}{Soiling Detection} & \multirow{1}{*}{ResNet10 \cite{he2016deep}} & Category (\%)   & 84.5           \\
Depth Estimation                   & Eigen \cite{eigen2015predicting}   & RMSE                         & 7.7            \\ 
Motion Segmentation            & MODNet \cite{siam2018modnet}                         & IoU                          & 45            \\ 

\multirow{2}{*}{Visual Odometry} & \multirow{2}{*}{ResNet50 \cite{he2016deep}} & Translation (\textless{}5mm)    &   51         \\
                                   &                                 & Rotation (\textless{}0.1\degree) &    71         \\


Visual SLAM           & LSD SLAM \cite{engel2014lsd}                          & Relocalization (\%)                          &  61             \\
\hline
\multicolumn{4}{c}{\textbf{3D Bounding Box Detection - Complex YOLO \cite{Simon2018ECCVWorkshops} }}\\ 
\textbf{Metric for Training}                      & \textbf{AP (\%)}                & \textbf{AOS (\%)}              & \textbf{Runtime (ms)} \\ \hline \hline
3D-IoU                      & 64.38                     & 85.60    & 95          \\ 
$S_{srt}$              & 62.46                     & 88.43   & 1           \\ 
\hline 
\end{tabular}

\label{tab:all-results}
\end{adjustbox}
\vspace{-0.5cm}
\end{table}

The task of soiling detection was to our best knowledge first defined in~\cite{Uricar-2019a}. 
Unlike the front camera which is behind the windshield, the surround view cameras are usually directly exposed to the adverse environmental conditions, and thus prone to becoming soiled or water drops forming on the lens. As the functionality of visual perception degrades significantly, detection of soiled cameras is necessary for achieving higher levels of automated driving. 
 As it is a novel task, we discuss it in more detail below.


We treat the camera soiling detection task as a mixed multilabel-categorical classification problem, i.e. we are interested in a classifier, which jointly classifies a single image with a binary indicator array, where each $0$ or $1$ corresponds to missing or present class, respectively and simultaneously assigns a categorical label. The classes to detect are $\{ \mathrm{opaque}, \mathrm{transparent} \}$. Typically, opaque soiling arises from mud and dust (Figure~\ref{fig:soiling} right image), and transparent soiling arises from water and ice (Figure~\ref{fig:soiling} left image). However, in practice it is common to see water producing ``opaque'' regions in the camera image.  

Annotation for $5$k images is performed by drawing polygons to separate soiled from unsoiled regions, so that it can be modeled as a segmentation task if necessary. We evaluate the soiling classifier's performance via an example-based accuracy measure for each task separately, i.e. the average Jaccard index of the testing set: $\frac{1}{n} \sum_{i=1}^{n} \frac{|\mathbf{Y}_i \cap \mathbf{Z}_i|}{|\mathbf{Y}_i \cup \mathbf{Z}_i|}$, where $\mathbf{Y}_i \in \mathcal{Y} = \{ 0, 1 \}^k$ denotes the label for the $i$-th testing sample, $\mathbf{Z}_i$ denotes the classifier's prediction and $n$ denotes the cardinality of the testing set and $k$ the length of the label vector.
We use a small baseline network (ResNet10 encoder + $3$-layer decoder) and achieved a precision of $84.5\%$ for the multilabel classification.

\subsection{3D Bounding Box Detection}

3D box annotation is provided for 10k frames with 3 classes namely `pedestrian', `vehicle' and `cyclist'. In general, 3D IoU~\cite{Geiger2012CVPR} is used to evaluate 3D bounding box predictions, but there are drawbacks, especially for rotated objects. Two boxes can reach a good 3D IoU score, while overlapping in total with an opposite heading. Additionally, an exact calculation in 3D space is a time consuming task. To avoid those problems, we introduce a new evaluation metric called Scaling-Rotation-Translation score (\textit{SRTs}). \textit{SRT} is based on the idea that two non-overlapping 3D boxes can easily be transformed with respect to each other by using independent rigid transformations: translation $S_t$, rotation $S_r$ and scaling $S_s$. Hence, $S_{srt}$ is composed by:
\vspace{-0.15cm}
\begin{equation*}
\begin{aligned}
S_s &= 1 - \min\Biggl(\frac{|1-s_x| + |1-s_y| + |1-s_z|}{w_s}, 1\Biggr) \\
S_r &= \max\Bigl(0, 1 - \frac{\theta}{w_r\pi}\Bigr) \ \
S_t = \max\Bigl(0, \frac{r_1+r_2-t}{r_1+r_2}\Bigr) \\
&\quad r_{1/2} = \frac{d_{1/2} \cdot w_{t}}{2} \quad \quad w_t, w_r, w_s \in (0, 1] 
\end{aligned}
\end{equation*}
\noindent where $s_{x,y,z}$ denotes size ratios in $x$, $y$, $z$ directions, $\theta$ determines the difference of the yaw angles and $t$ defines the Euclidean distance between the two box centers. $S_t$ is calculated with respect to the size of the two objects based on the length of the diagonals $d_{1/2}$ of both objects that are used to calculate two radii $r_{1/2}$. Based on the penalty term $p_t$ we define the full metric by:
\vspace{-0.15cm}
\begin{equation*}
\begin{aligned}
S_{srt} &=  p_t \cdot (\alpha\;S_s + \beta\;S_t + \gamma\;S_r) 
&\alpha + \beta + \gamma = 1\nonumber\\
p_t &= \begin{cases} 0,& \text{if } r_1 + r_2 < t\\
1,& \text{otherwise}
\end{cases}
\end{aligned}
\vspace{-0.35cm}
\end{equation*}
$w_s, w_t$ and $w_r$ can be used to prioritize individual properties (e.g. $w_s \rightarrow$ size, $w_t \rightarrow$ angle). For our baseline experiments we used $w_s = 0.3$, $w_t = 1$, $w_r = 0.5$, $\gamma = 0.4$ and $\alpha = \beta = 0.3$ to add more weight to the angle, because our experiments have shown that translation or scaling is easier to learn. 
For baseline, we trained Complex-YOLO \cite{Simon2018ECCVWorkshops} for a single class (`car'). We repeated training two times, first optimized on 3D-IoU \cite{Geiger2012CVPR} and second optimized on $S_{srt}$ using a fixed 50:50 split for training and validation. For comparison, we present 3D-IoU, orientation and runtime following ~\cite{Geiger2012CVPR} on \textit{moderate} difficulty, see Table ~\ref{tab:all-results}. Runtime is the average runtime of all box comparisons for each input during training. Even though this comparison uses 3D-IoU, we achieve similar performance for average precision (3D-IoU), with better angle orientation similarity (AOS) and much faster computation time.




\subsection{Monocular Depth Estimation}

Monocular Depth estimation is an important task for detecting generic obstacles.
We provide more than 100k images of all four cameras (totaling 400k) using ground truth provided by LiDAR.  Figure \ref{fig:teaser} shows a colored example where blue to red indicates the distance for the front camera. As the depth obtained is sparse, we also provide denser point cloud based on SLAM'd static scenes as shown in Figure \ref{fig:Velodyne}. The ground truth 3D points are projected onto the camera images using our proposed model discussed in Section \ref{sec:fisheyemodel}. We also apply occlusion correction to handle difference in perspective of LiDAR and camera similar to the method proposed in \cite{Kumar2018CVPRWorkshop}. We run the semi-supervised approach from \cite{Kumar2018CVPRWorkshop} using the model proposed by Eigen \cite{eigen2015predicting} as baseline on our much larger dataset and obtained an RMSE (Root Mean Square Error) value of 7.7. 

\subsection{Motion Segmentation}

In automotive, motion is a strong cue due to ego-motion of the cameras on the moving vehicle and dynamic objects around the vehicle are the critical interacting agents. Additionally, it is helpful to detect generic objects based on motion cues rather than appearance cues as there will always be rare objects like kangaroos or construction trucks. This has been explored in \cite{siam2018modnet,Valada_2017_IROS,siam2018rtseg} for narrow angle cameras. 
In our dataset, we provide motion masks annotation for moving classes such as vehicles, pedestrians and cyclists for over 10k images. We also provide previous and next images for exploring multi-stream models like MODNet \cite{siam2018modnet}. 
Motion segmentation is treated as a binary segmentation problem and IoU is used as the metric. 
Using MODNet as baseline network, we achieve an IoU of 45.

\begin{figure}[t]
    \centering
    \includegraphics[width=\columnwidth]{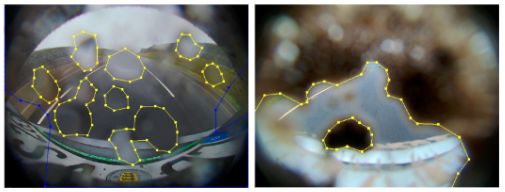}
    \vspace{-0.5cm}
    \caption{Soiling annotation}
    \label{fig:soiling}

\vspace{0.3cm}
    \centering
    \includegraphics[width=\columnwidth]{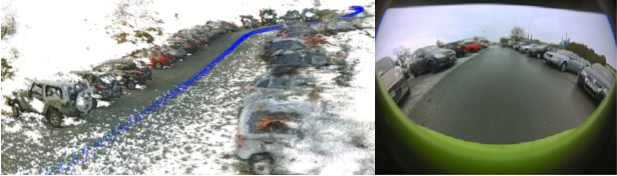}
    \vspace{-0.5cm}
    \caption{Visual SLAM baseline results (left) based on raw fisheye images (right)}
    \label{fig:Visual SLAM}
\vspace{0.3cm}

    \centering
    \includegraphics[width=\columnwidth,trim={0.5cm 5.5cm 0.5cm 0},clip]{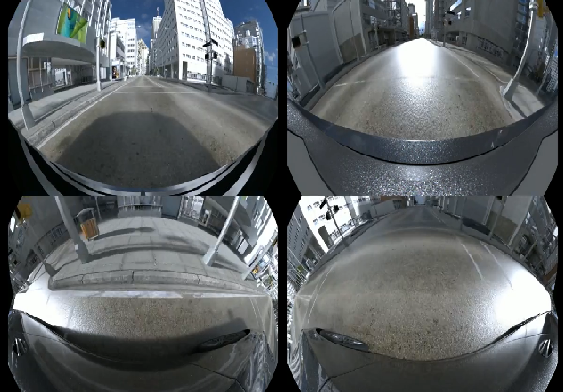}
    \vspace{-0.5cm}
    \caption{Synthetic images modelling fisheye optics}
    \label{fig:synthetic}
    \vspace{-0.5cm}
\end{figure}




\subsection{Visual Odometry/SLAM}

Visual Odometry (VO) is necessary for creating a map from the objects detected \cite{milz2018visual}. We make use of our GNSS and IMU to provide annotation in centimetre level accuracy. The ground truth contains all the six degrees of freedom upto scale and the metric used is percentage of frames within a tolerance level of translation and rotation error. Robustness could be added to the visual odometry by performing a joint estimation from all four cameras. 
We provide 50 video sequences comprising of over 100k frames with ground truth. The video sequences can also be used for Visual SLAM where we focus on relocalization of a mapped trajectory and the metric is same as VO. 
We use a fisheye adapted LSD-SLAM \cite{engel2014lsd} as our baseline model as illustrated in Figure \ref{fig:Visual SLAM} and accuracies are provided in Table \ref{tab:all-results}.

\subsection{Synthetic Data Domain Transfer}
Synthetic data is crucial for autonomous driving for many reasons. Firstly, it provides a mechanism to do rigorous corner case testing for diverse scenarios.
Secondly, there are legal restrictions like recording videos of a child. Finally, synthetic data is the only way to obtain dense depth and optical flow annotation. There are several popular synthetic datasets like SYNTHIA \cite{ros2016synthia} and CARLA \cite{Dosovitskiy17}. We will provide a synthetic version of our fisheye surround view dataset,
as shown in Figure \ref{fig:synthetic}. The main goal is to explore domain transfer from synthetic to real domain for semantic segmentation and depth estimation tasks. 

\subsection{End-to-End Steering/Braking}

Bojarski et al. demonstrated end-to-end learning \cite{bojarski2016end} for steering and recently it was applied to fisheye cameras \cite{toromanoff2018end}. Although this approach is currently not mature for deployment, it can be either used as a parallel model for redundancy or as an auxiliary task to improve accuracy of other tasks. 
In the traditional approach, perception is independently designed and it is probably a more complex intermediate problem to solve than what is needed for a small action space driving task. 
Thus we have added end-to-end steering and braking tasks to encourage modular end-to-end architectures and to explore optimized perception for the control task. The latter is analogous to hand-eye co-ordination of human drivers where perception is optimized for driving.

\section{Conclusions} \label{sec:conclusions}

In this paper, we provide an extensive multi-camera fisheye dataset for autonomous driving with annotation for nine tasks. We hope that the release of the dataset encourages development of native fisheye models instead of undistorting fisheye images and applying standard models. In case of deep learning algorithms, it can help understand whether spatial distortion can be learned or it has to be explicitly modeled. In future work, we plan to explore and compare various methods of undistortion and explicit incorporation of fisheye geometry in CNN models. We also plan to design a unified multi-task model for all the listed tasks.

\section*{Acknowledgement}

We would like to thank our colleagues Nivedita Tripathi, Mihai Ilie, Philippe Lafon, Marie Yahiaoui, Sugirtha Thayalan, Jose Luis Fernandez and Pantelis Ermilios for supporting the creation of the dataset, and to thank our partners MightyAI for providing high-quality semantic segmentation annotation services and Next Limit for providing synthetic data using \href{https://anyverse.ai}{ANYVERSE}  platform.

\newpage

{\small
\bibliographystyle{ieee_fullname}
\bibliography{bib/references}
}

\end{document}